\documentclass[preprint,review,12pt]{elsarticle}

\usepackage{amssymb}
\usepackage{doi}

\usepackage{amsmath}
\usepackage{changes}

\usepackage{amsfonts}
\usepackage{color,soul}
\usepackage{xcolor}
\usepackage{booktabs}
\usepackage{makecell}
\usepackage{algorithmic}
\usepackage{array}
\usepackage{url}
\usepackage{hyperref}

\hypersetup{
	colorlinks=true,
	linkcolor=black,
	filecolor=black,
	urlcolor=black,
	citecolor=black,
}

\newcommand\etal{\textit{et al.}}

\begin{document}

\begin{frontmatter}



\title{EGR-Net: A Novel Embedding Gramian Representation CNN for Intelligent Fault Diagnosis}


\author[1]{Linshan Jia} \ead{linshan.jia@my.cityu.edu.hk}

\address[1]{Department of Electrical Engineering, City University of Hong Kong, Hong Kong, SAR, 999077, China}

\begin{abstract}
Feature extraction is crucial in intelligent fault diagnosis of rotating machinery. It is easier for convolutional neural networks(CNNs) to visually recognize and learn fault features by converting the complicated one-dimensional (1D) vibrational signals into two-dimensional (2D) images with simple textures. However, the existing representation methods for encoding 1D signals as images have two main problems, including complicated computation and low separability. Meanwhile, the existing 2D-CNN fault diagnosis methods taking 2D images as the only inputs still suffer from the inevitable information loss because of the conversion process. Considering the above issues, this paper proposes a new 1D-to-2D conversion method called Embedding Gramian Representation (EGR), which is easy to calculate and shows good separability. In EGR, 1D signals are projected in the embedding space and the intrinsic periodicity of vibrational signals is captured enabling the faulty characteristics contained in raw signals to be uncovered. Second, aiming at the information loss problem of existing CNN models with the single input of converted images, a double-branch EGR-based CNN, called EGR-Net, is proposed to learn faulty features from both raw signal feature maps and their corresponding EGRs. The bridge connection is designed to improve the feature learning interaction between the two branches. Widely used open domain gearbox dataset and bearing dataset are used to verify the effectiveness and efficiency of the proposed methods. EGR-Net is compared with traditional and state-of-the-art approaches, and the results show that the proposed method can deliver enhanced performance.
\end{abstract}




\end{frontmatter}


\section{Introduction}\label{sec:intro}
Bearings and gearboxes are critical components of rotating machines\cite{lei2020applications}. These machines often operate under varying speeds, loads, material conditions, maintenance procedures, and environments. Thus, performing effective fault diagnosis for the equipment through vibrational signal analysis is challenging and has received significant attention\cite{mo2022conditional}.

Intelligent fault diagnosis based on deep learning (DL) has demonstrated improved performance on fault classification. Many DL models such as CNNs \cite{tang2022intelligent}\cite{li2022industrial}, generative adversarial networks(GANs) \cite{pan2021generative}, Deep Belief Networks (DBNs) \cite{gao2021rolling}, and transformers \cite{chen2022multi} are applied in fault diagnosis with promising results. Among those DL-based methods, the CNN model is developed to imitate the concept of visual human object recognition. CNN's feature extraction performance has been verified in many applications, such as image recognition \cite{krizhevsky2012imagenet} and video analysis \cite{xu2019spatiotemporal}. There are two main categories of CNN methods used in fault diagnosis depending on the dimensionality of the input data: two-dimensional CNN (2D-CNN) for processing 2D data like images\cite{shao2020intelligent}, and one-dimensional CNN (1D-CNN) for processing 1D data such as vibrational signals \cite{zhang2022selective}. For the 1D-CNN method, the 1D time series can be input into the model directly \cite{xu2021fault}. However, compared with 2D-CNN, 1D-CNN has certain disadvantages when applied to fault diagnosis. First, raw 1D vibrational signals are usually low separable, and learning features directly from the 1D signal is relatively inefficient \cite{wang2021intelligent}. Existing 1D-to-2D conversion methods in 2D-CNN models can improve the data separability and reduce redundancy. Second, the design of 2D-CNN is mature, and many well-established 2D-CNN models are available in many research literatures. Meanwhile, the input dimension of 2D-CNN is usually smaller than the 1D-CNN in fault diagnosis, making the 2D-CNN easier to be designed. Thus, 2D-CNN is studied in this paper.

For the 2D-CNN, the 1D vibrational signal is required to be converted into a 2D feature representation using various signal processing methods. The raw signal contains the richest faulty information, but irrelevant information in the raw signal often limits its separability. The function of the 1D-to-2D conversion is that it can be regarded as an initial feature extraction procedure that helps to enhance the convergence of the CNN model \cite{tang2020data}. The merits of the conversion processing include that the converted 2D features usually contain simpler textures, which are more representative than the raw signals. For example, Wen \etal \cite{wen2017new} transformed a 1D raw vibrational signal into the 2D signal matrix through the tensor reshaping method. The obtained signal matrices are then fed into the LeNet-5 model for bearing fault diagnosis. Similar to \cite{wen2017new}, Zare \etal \cite{zare2021simultaneous} proposed a signal-to-image conversion method for the wind turbine fault diagnosis. Shao \etal \cite{shao2018highly} employed the continuous wavelet transform as the input of the VGG-16 model with the transfer learning strategy. In \cite{pang2020investigation}, the bispectrum of raw vibrational signals is used as the input of transfer CNN to diagnose the planetary gearbox. In \cite{chen2020deep} the cyclic spectral coherence (CSC) connects with a LeNet model for bearing fault diagnosis because of its ability to exploit the second-order cyclo-stationary behavior of bearing vibration signals. Tang \etal \cite{tang2020novel} introduced Gramian angular field (GAF) as the 2D representation of the vibrational signals and then a 5-layer CNN model was designed to classify GAFs for the low-speed bearing fault diagnosis. Xiong \etal \cite{xiong2021application} proposed a similar Gramian matrix (SGM) based on dimensionless indices as the 2D representation input of the CNN for bearing fault diagnosis. Other representations for encoding time series as images, including synchro-extracting transform (SET)\cite{wang2021intelligent}, Markov transition field (MTF) \cite{han2021new}, and kurtogram \cite{saufi2020gearbox}, are also reported for the rotating machinery fault diagnosis.

Although the abovementioned 2D-CNN models have achieved promising results in various fault diagnosis tasks, there are still several problems that are not well solved. First, the existing representation methods for encoding 1D signals as images have two main problems, i.e., complicated computation and low separability. Specifically, the calculations of the existing 1D-to-2D conversion methods are usually complicated and inefficient. If the calculation of the conversion needs a complicated algorithm and consumes huge computational resources, then the conversion process may be unnecessary. Moreover, the complicated 1D-to-2D conversion calculation breaks the end-to-end nature of 2D-CNN. Furthermore, the 2D representation with low separability contributes little to the subsequent feature learning and fault recognition. Second, for those 2D-CNN fault diagnosis methods taking the converted 2D images as the only inputs, the information loss resulting from the conversion process is not well considered. Although the key feature in 2D representations may become more significant than the raw signals, certain critical information may be lost during the conversion. It is intriguing and meaningful to improve fault classification performance by reducing information loss.

Aiming at the above issues, this article first proposes a new 1D-to-2D conversion method called Embedding Gramian Representation (EGR), with easy computation and good separability. Unlike the existing Gramian-based method GAF \cite{tang2020novel} defined in the real space, our EGR conducts Gramian in the embedding space. Compared with the recently proposed SGM \cite{xiong2021application}, our EGR does not need to calculate various dimensionless indices and has a less computational burden. For a given time series in real space, it is first projected into a state vector sequence in the timeline in the embedding space. Then the temporal correlations between all pairs of state vectors in the vector sequence are calculated via inner product operation. The EGR method can be implemented in two simple steps: the construction of the raw signal matrix (RSM) and the calculation of the Gramian of the RSM. It is computational efficiently because the only calculation of the EGR algorithm is matrix multiplication. The generated features are separable well because the EGR only stores the temporal correlation information, and the information redundancy is reduced significantly. For the periodic vibrational signal, its EGR demonstrates stripe texture. The EGR has a clear physical meaning and explicit theoretical basis, and it can be easily embedded into the CNN model.

For the information loss problem of the commonly used 2D-CNN taking the converted 2D representation as the only input, a new double-branch 2D-CNN model called EGR-Net is designed to learn features from the RSM feature maps and EGRs simultaneously. In the EGR-Net, the RSM and EGR of the raw vibrational signal are processed by convolutional layers in parallel. The RSM contains original value information, and its corresponding EGR provides initial features with good separability. The proposed EGR-Net can reduce the information loss in existing models with a single input in an end-to-end manner. For making full use of the EGR, the bridge connection is innovatively introduced to improve the interactions between the two branches. Via the bridge connection, the feature maps of intermediate layers in the RSM branch are converted into EGRs, which are then concatenated in the EGR branch. The bridge connection design can further reduce information loss and strength the feature learning ability of the CNN model.

The main contributions of this paper are as follows:

(1) This paper proposes a new 1D-to-2D conversion method called EGR for capturing fault characteristics in raw vibrational signals. The proposed EGR is easy to calculate but can generate compact features with good separability. The EGR can reflect the temporal correlation information of the signal and unveil the fault-related characteristics. Meanwhile, the physical meaning of the EGR texture and the theoretical principle of the EGR are provided. The parameter selection rules are discussed.

(2) A double-branch CNN model termed as EGR-Net is proposed for reducing the information loss problem of the traditional CNNs with a single input. The RSM and EGR of the raw vibrational signal are the inputs of the two branches of the 2D-CNN model. Especially, the bridge connection is designed for converting the feature maps of the intermediate layers in the RSM branch into EGRs and feeding them into the EGR branch. It enhances the model feature representation ability remarkably.

(3) EGR-Net retains the end-to-end nature. Different from most existing 2D-CNN models for fault diagnosis, the operations for RSM and EGR in EGR-Net are only tensor reshaping and matrix multiplication, which can be performed inside the EGR-Net model. Thus, EGR-Net accepts inputs of 1D signals and gives outputs of fault classification results.

The rest of the article is organized as follows. The EGR and EGR-Net are described in section \ref{sec:methodology}. In section \ref{sec:exprimentvalid}, the proposed ECG-Net is applied to two datasets for verifying the merits. Comparisons with other traditional and state-of-the-art methods are also given in this section. Finally, the paper is concluded in section \ref{sec:conclusion}.

\section{Methodology}\label{sec:methodology}

In this section, aiming at the complicated computation and low separability problems of existing 2D representation methods, a new 1D-to-2D conversion method called EGR is introduced. In order to solve the information loss issue of conventional CNN models, based on the EGR method, a double-branch CNN model termed as EGR-Net is proposed, in which the RSM and EGR are inputs of the two branches of the 2D-CNN model. A bridge connection between the two branches is designed to further improve the model. In the following, the EGR and EGR-Net model will be provided in detail.

\subsection{Embedding Gramian Representation}
A new 1D-to-2D conversion method based on Gramian called EGR is proposed for the fault diagnosis. Considering that the time series $x=\{x(i)\},i=1,2,...,N$ with the length of $N=mn$ is the vibrational signal measured on a rotating machine. The signal $x$ can be regarded as the scalar sequence in the real space $\mathbb{R}$. It is natural to map the scalar sequence $x$ into a state vector sequence in the embedding space ${{\mathbb{R}}^{m}}$. Because there are $N=mn$ data points in $x$, the number of state vectors in the sequence should be $n$. According to the phase space reconstruction theory \cite{huffaker2017nonlinear}, the state vector sequence would be a discrete trajectory in embedding space \cite{broomhead1986extracting}, and it can represent how the dynamic system evolves in time. The $i$-th state vector of the vector sequence is defined as
\begin{equation}\label{Eq:01}
  {\boldsymbol{x}_i}={{[x(i),x(n+i),x(2n+i),...,x((m-1)n+i)]}^T}
\end{equation}
The vector sequence in the timeline can form a matrix:
\begin{align}\label{Eq:02}
   X& = [{{\boldsymbol{x}}_{1}},{{\boldsymbol{x}}_{2}},...,{{\boldsymbol{x}}_{n}}] \notag \\
    & = \begin{bmatrix}
          x(1) & x(2) & \cdots  & x(n)  \\
          x(n+1) & x(n+2) & \cdots  & x(2n)  \\
          x(2n+1) & x(2n+2) & \cdots  & x(3n)  \\
          \vdots  & \vdots  & \ddots  & \vdots   \\
          x(mn-n+1) & x(mn-n+2) & \cdots  & x(mn)  \\
        \end{bmatrix}
\end{align}

\begin{figure}[!t]
  \centering
  \includegraphics[scale=0.3]{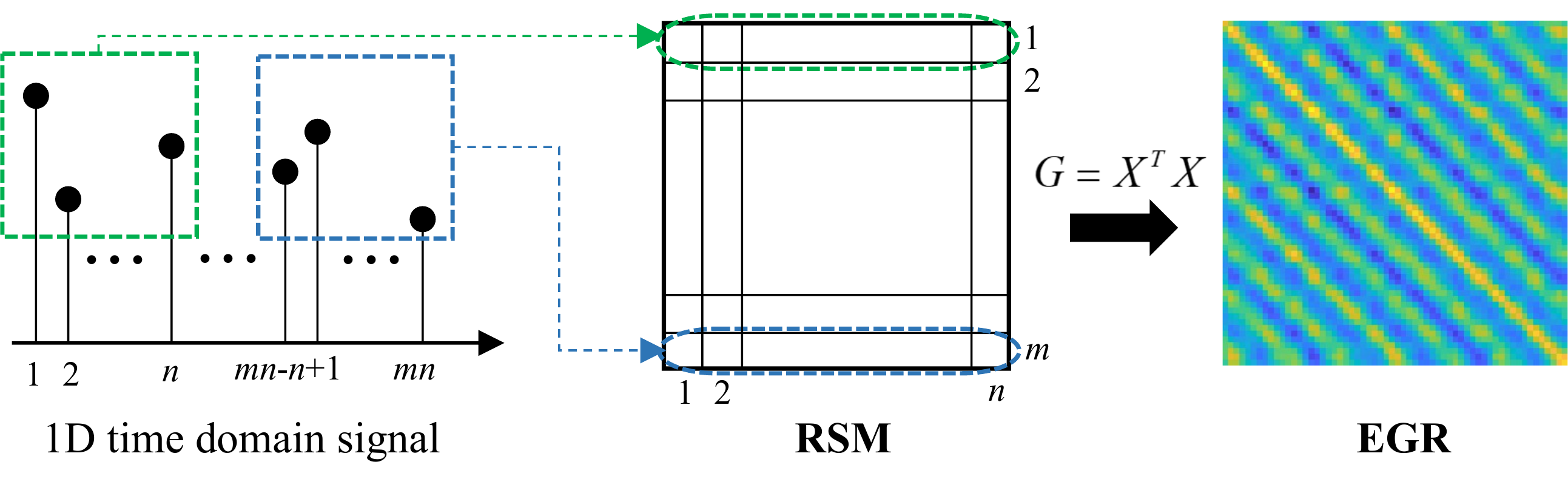} 
  \caption{The diagram of EGR.}
  \label{fig:EGR}
\end{figure}

The matrix $X$ contains all the original datapoints of the signal $x$, and we call the matrix $X$ as the raw signal matrix (RSM). The construction of matrix $X$ is similar to \cite{wen2017new}. Wang \etal \cite{wang2015imaging} proposed a time series representation method named Gramian angular field (GAF), in which the temporal correlation within different time intervals is identified. Inspired by this work, Gramian is adopted to calculate the temporal correlation between all pairs of state vectors in the RSM $X$. The obtained temporal correlation can reflect the dynamic characteristics of the signal $x$, and the Gramian $G\in {{\mathbb{R}}^{n\times n}}$ is expressed as
\begin{equation}\label{Eq:03}
  G =\text{Gram}(X) = X^{T}X.
\end{equation}
The symmetric matrix $G$ is actually the EGR of signal $x$.
Therefore, as shown in Fig. \ref{fig:EGR}, the proposed EGR contains two steps, i.e., construction of the raw signal matrix and calculation of Gramian of column vectors of the raw signal matrix. To the best of our knowledge, this paper is the first literature that takes the Gramian of the state vectors as the 2D representation of the vibrational signals.

When bearings become faulty, the high-frequency resonance is evoked by the generated impulses, and the fault characteristics can be reflected in the periodic impulses \cite{liu2016time}. For gearbox fault, a faulty point will also excite structural resonances and show substantial periodic property \cite{antoni2002differential}. One question concerned about is what the EGR would demonstrate for periodic vibrational signals. To answer this question, the first thing needed to clarify the physical meaning of each element in $G\in {{\mathbb{R}}^{n\times n}}$. The EGR consists of scalar products of all pairs of points on the trajectory in the embedding space, and it reflects the intrinsic structure of the trajectory. The $G$ is a symmetric matrix, and each of its diagonal describes the correlation between the two state vectors with the same time interval. The $k$-th diagonal of $G$ is calculated as the inner product of ${{\boldsymbol{x}}_{i}}$ and ${{\boldsymbol{x}}_{i+k}}$, that is ${{\boldsymbol{x}}_{i}}^{T}{{\boldsymbol{x}}_{i+k}}$. For the main diagonal, it corresponds to $k=0$ and it would be ${{\boldsymbol{x}}_{i}}^{T}{{\boldsymbol{x}}_{i}}$.

After the physical meaning of $G$ is explicated, the texture of $G$ for periodic vibrational signals is then discussed. If the time series $x=\{x(i)\},i=1,2,...,N$ has the period of $T$ in real space $\mathbb{R}$, then $x(i)=x(i+T)$ is derived. Through downsampling, it can be easily derived that $x(in)=x(in+T)$. Then, $x(in+j)=x(in+j+T)$ is obtained. Therefore, the state vectors would satisfy ${{\boldsymbol{x}}_{j}}={{\boldsymbol{x}}_{j+T}}$. That is to say, if the time series $x$ is periodic in $\mathbb{R}$, then its state vector sequence is also periodic in the timeline in embedding space ${{\mathbb{R}}^{m}}$. As mentioned before, the $k$-th diagonal of the EGR representation is ${{\boldsymbol{x}}_{i}}^{T}{{\boldsymbol{x}}_{i+k}}$, the values on diagonals would be larger when $k=T$. Theoretically, the texture of EGR of $x$ would demonstrate the light and dark stripe textures on the diagonals, as shown in Fig. \ref{fig:EGR}. This kind of texture generally is simpler than the original raw signal. According to our experiments, the dominated periodic frequency in EGR is consistent with the frequency spectrum peak with the highest power. Thus, it can be said that the 2D feature representation using EGR is to capture the periodic dynamic fluctuations in signals.

The other question concerned is how to determine the embedding space dimension $m$ and the number $n$ of state vectors in the vector sequence. For a discrete signal with fixed length $N=mn$, obviously, there is a trade-off between $m$ and $n$. Actually, there is no specific rule for the two parameters, but some guidance can be given to determine them qualitatively. For the space dimension $m$, if $m$ is small, the degree of statistical independence between state vectors in $X$ would decrease. The drop in statistical independence would lower the quality of the correlation information in ECG representation. However, if $m$ is too large, then the state number $n$ would be very small, which would lead to little correlation information that can be extracted by EGR.

As for the determination of the state vector number $n$, it is highly related to the vibrational signal itself. Let ${{f}_{s}}$ denote the sampling frequency of signal $x$. It has been discussed that the texture of EGR should be stripes with period $T$ on diagonals. Actually, the period $T$ is limited by the EGR's shape, which is $n\times n$. It can be directly seen that the effective range of periodic frequencies that ECR can reflect is from ${{{f}_{s}}}/{n}$ to ${{f}_{s}}$. The low periodic frequency below ${{{f}_{s}}}/{n}\;$ would not be illustrated as stripe texture in EGR representation. The larger $n$ can enlarge the EGR's frequency range. Fortunately, for the vibrational signals in the real world, the key information related to the fault is located in the high-frequency band. On some special occasions, the sampling frequency may be very high; it can be solved by downsampling the signal to ensure that there are enough stripes in the EGR. Experimentally, the data length $N$ of the signal sample is suggested to select the even integral power of an integral number ranging from 1000 to 10000, e.g., $N={{64}^{2}}=4096$. Then the $m$ and $n$ can be determined as the square root of the data length, that is $m=n=64$.

The proposed EGR has four main important properties. First, the algorithm of EGR is simple and can be understood easily. The only calculation involved in EGR is matrix multiplication, which is computationally cheap. Second, the EGR can capture the intrinsic periodic fluctuation. The extracted features are fault-related and would show good separability. Third, compared with the raw signal, the information redundancy in the EGR is reduced significantly. It means that the feature would be compact. Finally, the simple computation of EGR makes it easy to combine EGR with the CNN model without breaking its end-to-end nature.

\subsection{EGR-Net}
In 2D-CNN fault diagnosis methods, the 1D time-series signal needs to be transformed into a 2D feature map. Except for the special cases of RSM (i.e., \cite{wen2017new}), most 1D-to-2D methods based on complicated signal processing algorithms improve feature representative ability at the cost of inevitable information loss. The recognition accuracy of the CNN model with the single input of 2D representation would decrease if the lost information contained the effective fault features.

\begin{figure}[!t]
  \centering
  \includegraphics[scale=0.15]{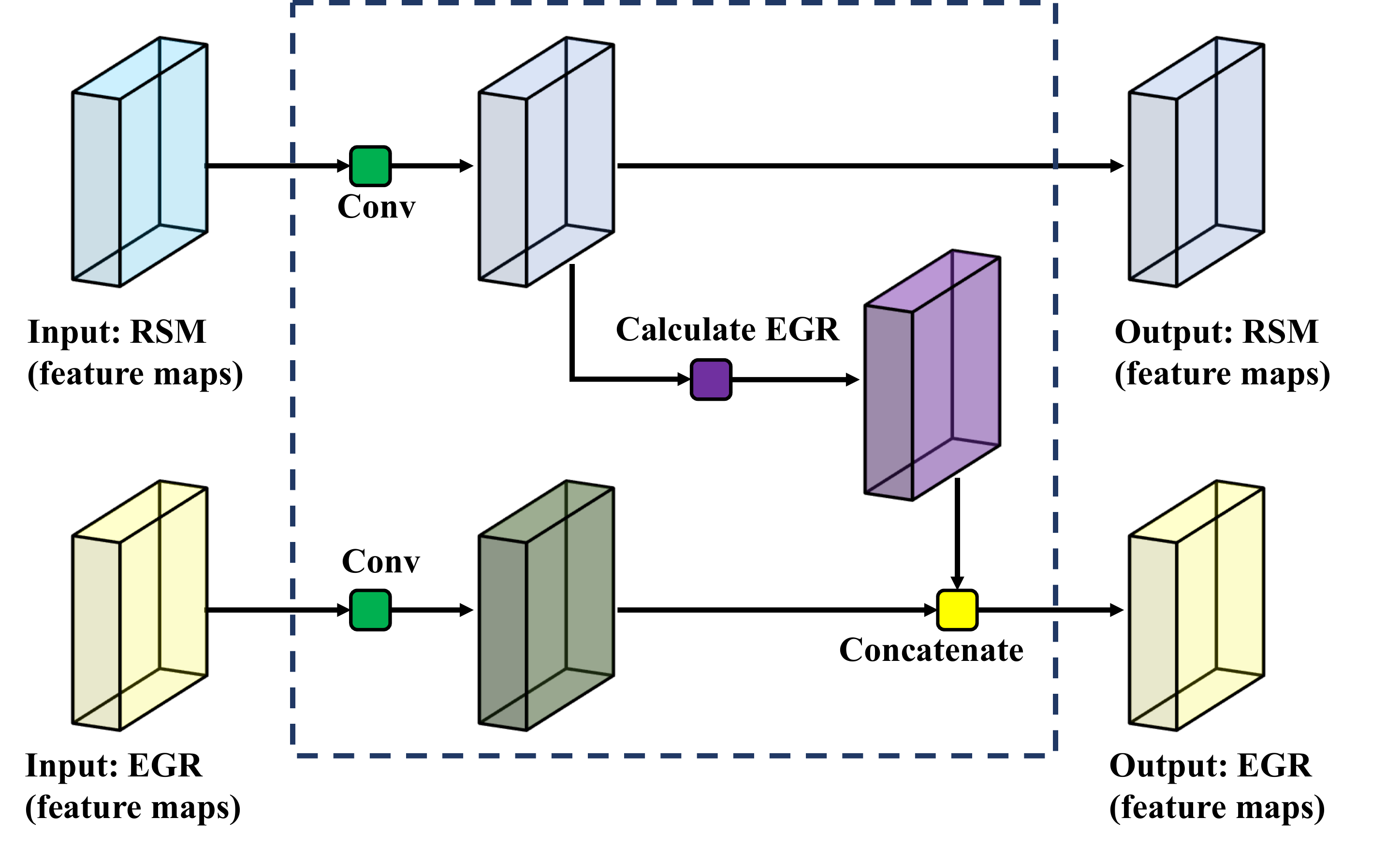}
  \caption{The basic Gramian convolutional block (GCB) of the proposed EGR-Net Model.}
  \label{fig:GCB}
\end{figure}

To address this issue, the EGR-Net is presented. The basic Gramian convolutional block (GCB) of EGR-Net is shown in Fig.~\ref{fig:GCB}. The 1D signal $x$ is first converted into RSM $X$ and EGR $G$. Then, a convolutional layer is used to obtain the corresponding feature maps. To mitigate the gradient diffusion and improve the generalization, batch normalization (BN) \cite{ioffe2015batch} is adopted to reparametrize feature maps adaptively. The feature maps after BN cascade the activation layer. Rectified linear unit (ReLU) \cite{nair2010rectified} is used to avoid saturation and gradient vanishing compared with other activation functions such as sigmoid and tanh. The output of the convolutional layer can be expressed as
\begin{equation}\label{Eq:04}
  \begin{split}
    & {{X}_{h}}=\text{ReLU}(\text{BN}(w\otimes X+b)) \\
    & {{G}_{h}}=\text{ReLU}(\text{BN}(w\otimes G+b)), \\
  \end{split}
\end{equation}
where $\otimes$ denotes the convolution operation, $w$ represents the kernels, $b$ is the bias, BN is the batch normalization, and ReLU is the ReLU activation function.

Next, the EGRs of feature maps ${{X}_{h}}$ of RSM $X$ are calculated and then normalized using layer normalization (LN) \cite{ba2016layer}, it can be written as
\begin{equation}\label{Eq:05}
  X{{G}_{h}}=\text{LN}(\text{Gram}({{X}_{h}})).
\end{equation}
The $X{{G}_{h}}$ is then concatenated to the feature maps ${{G}_{h}}$ of EGR $G$ through channels as the block output of the EGR branch. Meanwhile, the feature maps ${{X}_{h}}$ are the block output of the RSM branch. The outputs of the block can be written as
\begin{equation}\label{Eq:06}
  \begin{split}
    & {{X}_{o}}={{X}_{h}} \\
    & {{G}_{o}}=\text{Concat}([{{G}_{h}},X{{G}_{h}}]). \\
  \end{split}
\end{equation}
The Eq. (\ref{Eq:05}-\ref{Eq:06}) describe the bridge connection between the two branches. The EGR-Net can learn features not only from the EGR of the raw signal, but also from the feature maps of intermediate layers of the RSM branch.

The EGR-Net can be obtained by stacking GCBs. Supposed that there are $l$ GCBs in the network, and let the outputs of a GCB be
\begin{equation}\label{Eq:07}
\left\{ {{X}_{o}},{{G}_{o}} \right\}=\text{GCB}\left( \left\{ X,G \right\} \right).
\end{equation}
Thus, the outputs of the last GCB are written as
\begin{equation}\label{Eq:08}
\left\{ {{X}_{o}}^{l},{{G}_{o}}^{l} \right\}=\text{GCB}\left( \text{GCB}...\text{GCB}\left( \left\{ X,G \right\} \right) \right).
\end{equation}
The learned feature maps (i.e., ${{X}_{o}}^{l}$ and ${{G}_{o}}^{l}$) from the RSM branch and EGR branch are concatenated into a united feature map set through channels, and it is expressed as
\begin{equation}\label{Eq:09}
O=\text{Concat}([{{X}_{o}}^{l},{{G}_{o}}^{l}]).
\end{equation}

In order to mitigate overfitting and improve generalization \cite{peng2020multibranch}, global average pooling (GAP) is used for obtaining the final feature vector $y$, that is
\begin{equation}\label{Eq:10}
y=\text{GAP}(O).
\end{equation}
Followed by a fully connected layer with SoftMax as the classifier, the fault classification results can be derived by finding the index with the maximum probability. It is expressed as
\begin{equation}\label{Eq:11}
  \begin{split}
  & {{Q}_{j}}=w{{y}^{j}}+b \\
  & {{P}_{j}}=\frac{\exp ({{Q}_{j}})}{\sum\limits_{i}{\exp ({{Q}_{i}})}}, \\
\end{split}
\end{equation}
where $w$ and $b$ are the weight matrix and bias, respectively. ${{y}^{j}}$ denotes feature vector of the ${{j}_{th}}$ class, ${{P}_{j}}$ represents the possibility of the input $x$ belonging to the class $j$. The cross-entropy (CE) loss between predicted values and ground truth values in training data can be calculated as
\begin{equation}\label{Eq:12}
Loss=-\frac{1}{N}\sum\limits_{j=1}^{N}{\hat{{P}_{j}}\log {{P}}_{j}},
\end{equation}
where $N$ is the number of samples,  and ${{\hat{P}}_{j}}$ is the ground truth. The gradient of loss can be calculated and backpropagated to update the parameters in the whole model.


Compared with the existing CNN models, EGR-Net has the following characteristics. First, the double-branch structure ensures that the network learns from the highly separable features without information loss. The first branch is designed to extract original value information from RSM without information loss. The second branch is responsible for extracting effective information from the EGR. The EGR is more separable than RSM and can boost the convergence in the model training. Second, the bridge connection linking the two branches ensures that all the EGRs of the RSM feature maps in the intermediate layers can be used for feature extraction. The EGR feature maps and the generated EGRs from RSM branch enhance the feature fusion.

\subsection{The Structure of the EGR-Net}

The detailed structure of the EGR-Net is provided in Table.~\ref{Tab:EGR_Net_Structure}. From the table, it can be seen that there are a total of seven layers in the model, including five GCBs, one GAP for feature vector generation, and one SoftMax layer as the classifier. For the GCB layer, the kernel size, channel, and stride of its convolutional operation are the main concerned hyperparameters. The kernel sizes are set as 5, and the channels are set as 32 in the first two GCB layers. The reason why large kernel size and small kernel number are used is that firstly the feature map in the bottom layer usually has a larger size, which needs a larger local receptive field for the feature learning. Second, a small kernel number will balance the increased computational complexity due to the large kernel size. The third to fifth GCB layers adapt kernels with the size of 3. The kernel numbers of the third and fourth convolutional layers are 64, and the fifth is 128. It is because the feature maps become smaller when the network goes deeper, and more kernels can ensure that more valuable features can be extracted. It should be noted that the model takes inputs of 1D signal in practical implementation. The reshaping and matrix multiplication calculations of RSM and EGR are processed inside the 2D-CNN model. Thus, the EGR-Net is an end-to-end model.

\begin{table}[!t]
  \centering
  \caption{Network structure of the EGR-Net}
  \label{Tab:EGR_Net_Structure}
  \begin{tabular}{lllll}
      \toprule
      Layer & Kernel & Channel & Stride & Layer type \\
      \midrule
      0 & - & - & - & Input Layer \\
      1 & 5 & 32 & 1 & GCB \\
      2 & 5 & 32 & 1 & GCB \\
      3 & 3 & 64 & 1 & GCB \\
      4 & 3 & 64 & 2 & GCB \\
      5 & 3 & 128 & 2 & GCB \\
      6 & - & - & - & GAP \\
      7 & - & - & - & SoftMax \\
      \bottomrule
  \end{tabular}
\end{table}

\subsection{The Overall Framework of the Proposed Method}

In this paper, an intelligent fault diagnosis method for rotating machinery is proposed. The flowchart of the proposed method is given in Fig.~\ref{fig:Flowchart_of_Framework}. The steps of the framework are summarized as follows:

\begin{figure*}[!t]
  \centering
  \includegraphics[scale=0.22]{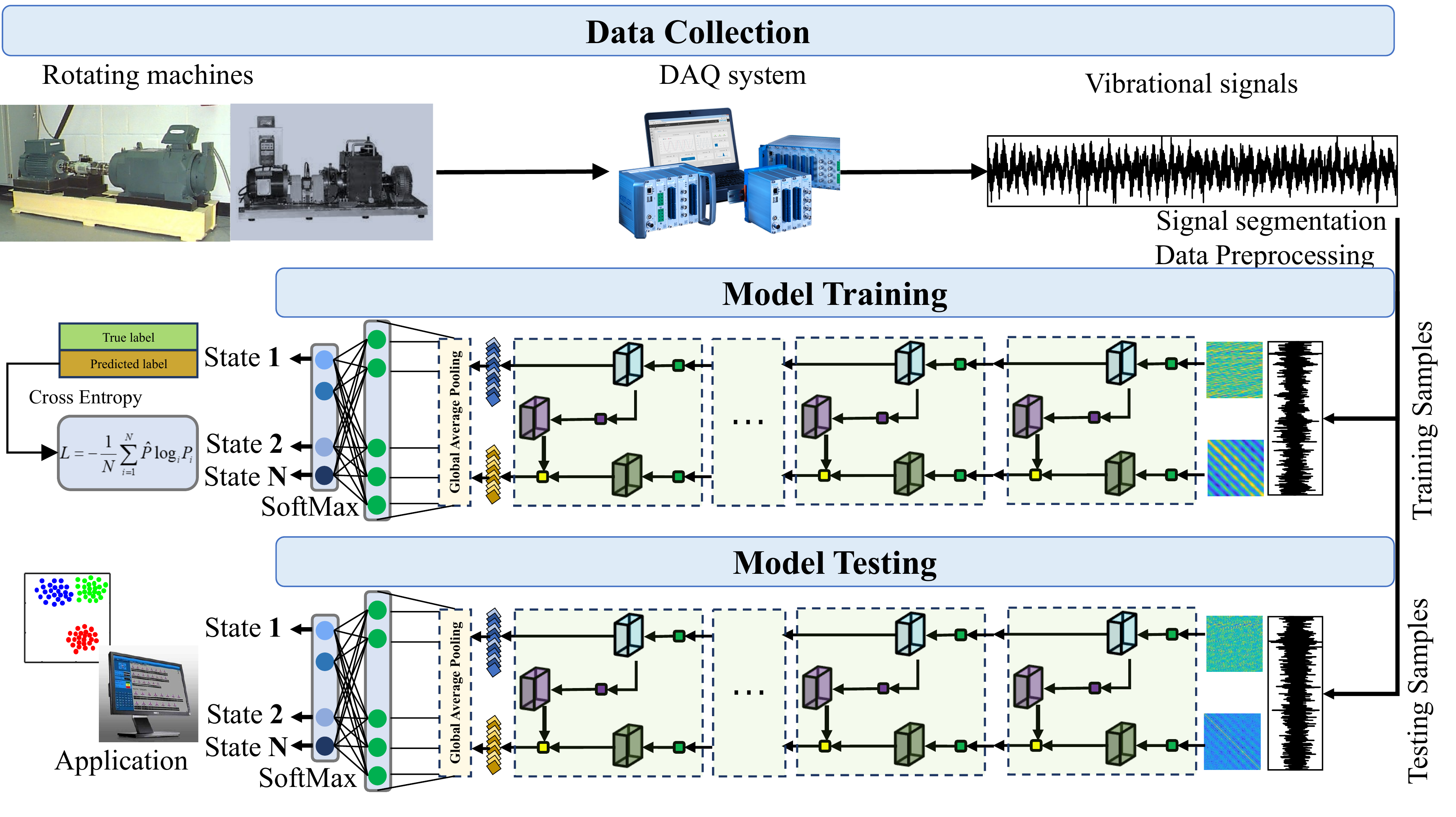} 
  \caption{The flowchart of the proposed method.}
  \label{fig:Flowchart_of_Framework}
\end{figure*}

Step 1: Data collection. Mount the accelerometer near the key components (i.e., bearings or gears) and record vibrational signals from the sensors using the data acquisition (DAQ) system.

Step 2: Signal segmentation and data preprocessing. Segment the raw vibrational signal into a series of samples containing specific data points. Divide the signal samples into the training set and the test set.

Step 3: Data conversion. Normalize the training samples and convert samples into RSMs and EGRs successively.

Step 4: Training the EGR-Net model offline. Feed the RSMs and EGRs into the EGR-Net model for training. The CE loss between ground truth and predicted label is calculated and minimized by the gradient descent optimization method, and the model weights are updated by the backpropagation algorithm.

Step 5: Fault diagnosis online. The new vibrational signals from the machine are processed the same as the samples during training. The output probability distribution of the deployed trained model gives the health states of the monitored equipment.

\section{Experimental Validation and Discussion}\label{sec:exprimentvalid}
In this section, the experimental setting is first given, then a widely used gearbox dataset and a bearing dataset are used to verify the effectiveness and efficiency of the EGR and the EGR-Net model.

\subsection{Experimental Setup}
The proposed EGR-Net is implemented using Python 3.6 based on TensorFlow 2.4.1 platform. The training and testing processes of the model are run on RedHat 4.8.5 operating system with GeForce RTX 2080Ti GPU. The raw signal matrix is normalized using the method proposed in \cite{wang2019understanding}, in which the mean value is first subtracted from the raw signal, then the zero-mean result is divided by the variance. The loss function in the training process is cross-entropy. Adam algorithm \cite{kingma2014adam} is selected as the optimizer. The batch size is set as 32, and the model runs 50 epochs. In order to accelerate the convergence, a large learning rate is used in the beginning stage of the training, and then we reduce it. Here, the initial learning rate is set as 0.0001, then the learning rate decrease to 0.1 times of the previous one every 15 epochs. We will make the source code public for the convenience of reproducing our results by other researchers.

Classification accuracy is the commonly used indicator of reflecting the performance of a model. In this paper, the accuracy is defined as
\begin{equation}\label{Eq:13}
  \text{Accuracy}= \frac{TP+TN}{TP+TN+FN+FP} \times 100\%,
\end{equation}
where $TP$, $TN$, $FN$, and $FP$ are the number of samples which are true positive, true negative, false negative and false positive, respectively\cite{wang2019understanding}.

The test rig can simulate the faulty characteristic of machines, but the vibrational signals collected from the real-world machines usually work in harsh environments on the strong noise background. Therefore, the white Gaussian noises with different signal-to-noise ratios (SNRs) are added into the raw vibrational signals to simulate the noise disturbance. The SNR (dB) is defined as
\begin{equation}\label{Eq:14}
  \text{SNR}=10{{\log }_{10}}\left( \frac{{{P}_{signal}}}{{{P}_{noise}}} \right),
\end{equation}
where ${{P}_{signal}}$ and ${{P}_{noise}}$ are the power of the signal and the noise, respectively\cite{wang2019understanding}.

\subsection{Case Study 1: Fault Diagnosis of Sun Gear Unit}
\subsubsection{Dataset Description}
\begin{figure}[!t]
  \centering
  \includegraphics[scale=0.21]{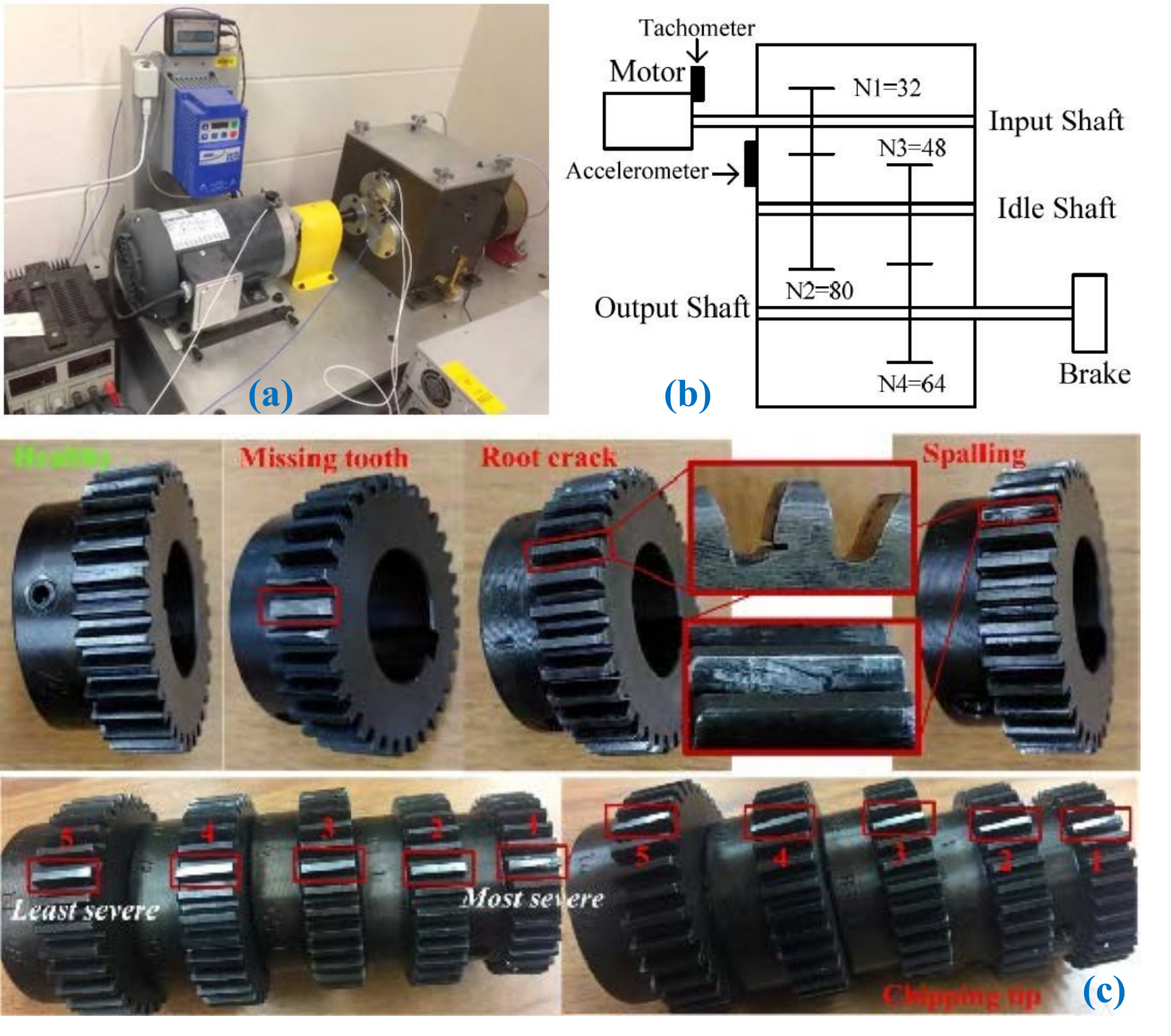}
  \caption{The structure of the UCONN test rig and faulty gears (a) general view of the test rig; (b) The schematic diagram of the gear system; (c) the faulty gears of nine fault types. \cite{cao2018preprocessing}}
  \label{fig:UCONN_test_rig}
\end{figure}

The gearbox fault dataset is provided by Tang \etal \cite{cao2018preprocessing} from University of Connecticut (UCONN). This dataset has been adopted by many published works for experimental verification \cite{ruan2021relation}\cite{shao2021modified}. As shown in Fig. \ref{fig:UCONN_test_rig}, nine fault types, including missing tooth, root crack, spalling, and chipping tip with five damage degrees, were seeded in the input shaft pinion (tooth=32). The different fault types of the dataset are summarized in Table. \ref{Tab:UCONN_Dataset}. The vibrational signals measured by the accelerometer are recorded under the sampling frequency of 20kHz.

\begin{figure}[!t]
  \centering
  \includegraphics[scale=0.2]{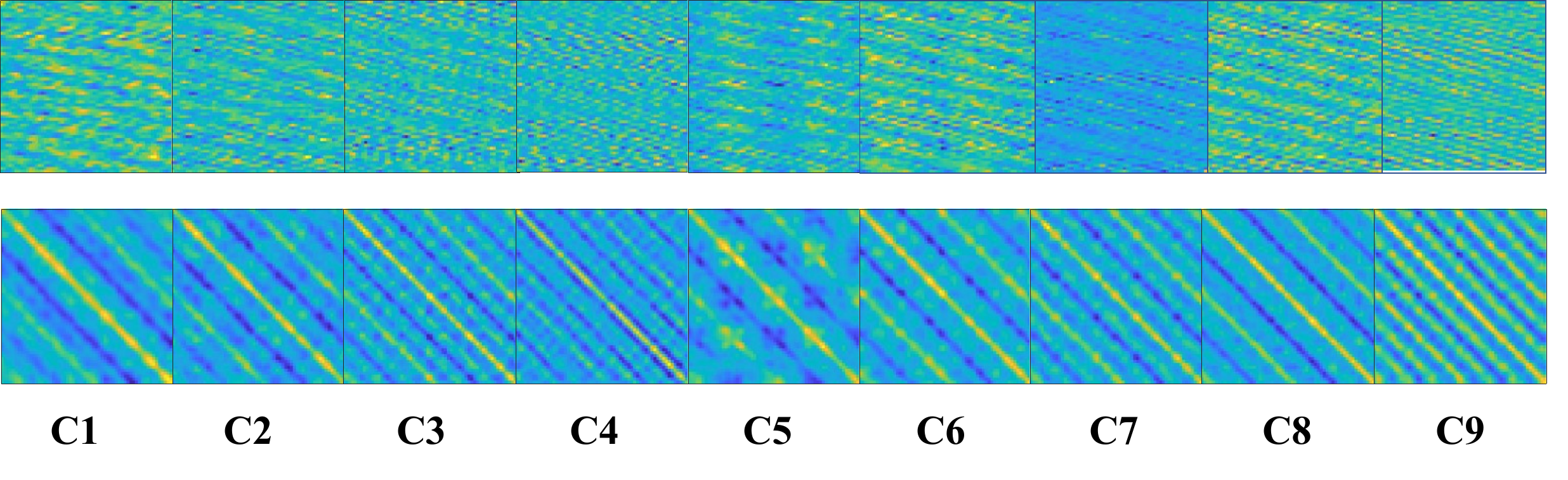}
  \caption{The RSMs (first row) and EGRs (second row) of nine health conditions of the UCONN gearbox dataset.}
  \label{fig:UCONN_RSN_EGR}
\end{figure}

In this dataset, each health condition contains 104 signals with the length of 3600. The size of RSM is set as $60\times 60$, and the size of EGR should therefore be $60\times 60$. There are $104\times9=936$ samples in total. Moreover, the first 50\% of the samples (i.e., 52 samples for each health condition) are used for training, and the rest 50\% are used for testing. The raw signal matrices and EGRs of nine health conditions are illustrated in Fig. \ref{fig:UCONN_RSN_EGR}. The textures of RSMs are relatively irregular. However, after the EGR conversion, the differences between different fault types become clear. The light and dark stripe textures also appear in the EGRs of gear vibrational signals. For different kinds of gear faults with Several severe degrees, the modulated frequencies and amplitudes in raw signals vary, which leads to different patterns in EGRs. Compared with the raw signal matrices, EGRs can provide simpler and more significant representations, which will raise the subsequent feature learning.

\begin{table}[!t]
  \centering
  \caption{Description of nine gear conditions}
  \label{Tab:UCONN_Dataset}
  \begin{tabular}{lllc}
  \toprule
  Gear conditions & Severity & Label \\
  \midrule
  Healthy          & Null             & C1 \\
  Missing tooth    & Null             & C2 \\
  Root crack       & Null             & C3 \\
  Spalling         & Null             & C4 \\
  Chipping tip5    & 5 (least severe) & C5 \\
  Chipping tip4    & 4                & C6 \\
  Chipping tip3    & 3                & C7 \\
  Chipping tip2    & 2                & C8 \\
  Chipping tip1    & 1 (most severe)  & C9 \\
  \bottomrule
  \end{tabular}

\end{table}

\subsubsection{Separability of EGR on Gear Fault Diagnosis}
In order to evaluate the effectiveness of the EGR on fault diagnosis, the separability of the EGR under background noise (SNR=0dB) is investigated. Six 1D-to-2D conversion methods adopted in recent fault diagnosis works, including kurtogram \cite{saufi2020gearbox}, bispectrum \cite{pang2020investigation}, cyclic spectral coherence (CSC) \cite{chen2020deep}, Gramian Angular Summation Field (GASF), Gramian Angular Difference Field (GADF) \cite{tang2020novel}, and Markov transition field (MTF) \cite{han2021new} are used for the comparative analysis. The separability of the EGR is verified and visualized by t-SNE \cite{van2008visualizing}. The raw vibrational signals (SNR=0dB) are converted into 2D representation by the seven methods and then visualized by t-SNE directly.

\begin{figure*}[!t]
  \centering
  \includegraphics[scale=0.15]{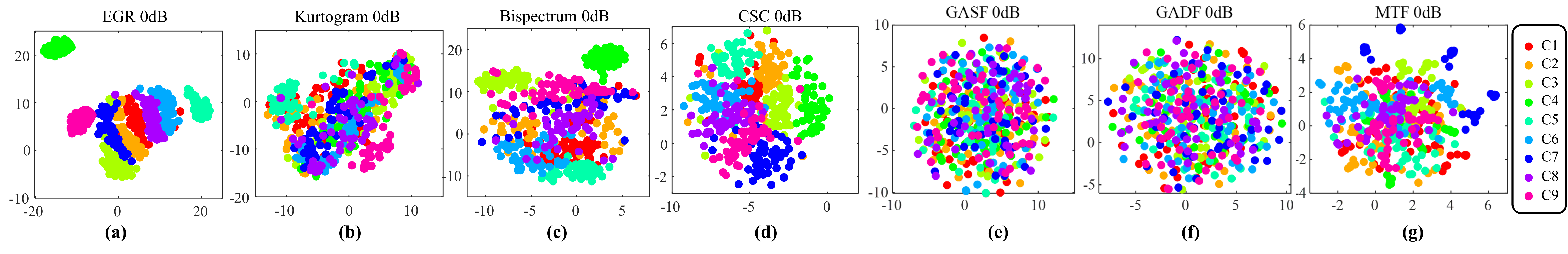}
  \caption{The t-SNE of the seven 1D-to-2D methods under SNR=0dB. (a) EGR; (b)  Kurtogram; (c) Bispectrum; (d) CSC; (e) GASF; (f) GADF; (g) MTF.}
  \label{fig:UCONN_TSNE_7_METHODS}
\end{figure*}

The t-SNE visualizations of different 2D representations of the test set are shown in Fig.~\ref{fig:UCONN_TSNE_7_METHODS}. The figure shows that the features of the introduced EGR are clustered with the best compactness, and the nine classes have the clearest boundaries in all seven 1D-to-2D conversion methods. However, the separation of the kurtogram between the fault classes is not good because different classes overlap severely. Meanwhile, bispectrum can discriminate class 3, class 4, and class 5 from the nine classes, but other classes show relatively dispersed. The CSC's separability is better than bispectrum and kurtogram, and it can distinguish most classes except class 1 (health condition). Nevertheless, the concentration degree of CSC is much lower than EGR. The GASF, GADF, and the MTF are proposed by Wang \etal \cite{wang2015imaging}. Although the GASF and the GADF are also based on Gramian, the Fig.~\ref{fig:UCONN_TSNE_7_METHODS} (e) and (f) show that the generated features are inseparable at all. The clustering of MTF looks better than GASF and GADF, but it is still not as good as CSC and our proposed EGR.

Therefore, it can be concluded that the introduced EGR achieves better discrimination of features compared with other commonly used 2D representations. With the separable features as the model input, the CNN model would have more capacity to focus on learning high-quality features from the vibrational signal. The features with good separability brought by EGR would also make model training easier and improve classification performance significantly.

\subsubsection{Fault Diagnosis Results Analysis and Visualization}

In order to assess the fault diagnosis performance of the proposed EGR-Net, experiments under four different SNRs (i.e., -6dB, -4dB, -2dB, and 0dB) are conducted. To reduce randomness from noise adding and model training, five trials for each SNR situation are run, and detailed classification accuracy (\%) results on the test set are shown in Table.~\ref{Tab:UCONN_Results_EGR_Net_five_runs}. For each run, the accuracy of the classification is calculated as Eq. (\ref{Eq:13}). The mean value and standard deviation of the testing accuracy are also given. As listed in Table.~\ref{Tab:UCONN_Results_EGR_Net_five_runs}, the average accuracy decreases with the increase of noise power. The EGR-Net achieves the accuracy of as high as 100.00\% under SNR=0dB, but it encounters a drop of 2.91\% when the SNR becomes -6dB. And also, the larger noise strength leads to larger standard deviation, which indicates that the noise would interfere with the classification results.

\begin{table}[!t]
  \centering
  \caption{Classification accuracy (\%) results of EGR-Net under four SNRs (running five times)}
  \label{Tab:UCONN_Results_EGR_Net_five_runs}
  \begin{tabular}{ccccc}
    \toprule
    SNR & -6dB & -4dB & -2dB & 0dB \\
    \midrule
    ~ & 96.37 & 98.29 & 99.15 & 100.00 \\
    ~ & 98.29 & 99.15 & 100.00 & 100.00 \\
    ~ & 95.94 & 97.65 & 99.79 & 100.00 \\
    ~ & 98.08 & 99.79 & 100.00 & 100.00 \\
    ~ & 96.79 & 98.93 & 99.57 & 100.00 \\
    \midrule
    Mean & \textbf{97.09} & \textbf{98.76} & \textbf{99.70} & \textbf{100.00} \\
    \midrule
    Std & 1.04 & 0.82 & 0.36 & 0.00 \\
    \bottomrule
  \end{tabular}

\end{table}

Although the proposed EGR-Net shows great fault diagnosis results, the computational performance should also be concerned. In the following, the complexity of EGR and EGR-Net are analyzed, respectively. First, All the calculation the EGR involves is the matrix multiplication. In EGR, ${{X}^{T}}\in {{\mathbb{R}}^{n\times m}}$ is multiplied by $X\in {{\mathbb{R}}^{m\times n}}$, the computational complexity is $\mathcal{O}({{n}^{2}}m)$. In modern computers, the calculation of matrix multiplication is highly optimized and can be processed rapidly. Second, the fault diagnosis system is often deployed online in real-world scenarios, and the real-time performance of the total system is really important. In deep learning, FLOPs are often used to describe how many operations are required to run a single instance of a given model. In this case, the FLOPs of EGR-Net are only 1.26 billion. Compared with classical CNN models like ResNet18 (1.8 billion) and VGG16 (15.3 billion), our proposed EGR-Net is a small network. According to our experimental study, the inference time for an instance is 5.82ms, which is much smaller than the duration time for a sample ($3600/20\text{kHz}=180\text{ms}$). Thus, the trained model can meet real-time requirements for online deployment.

\subsubsection{Ablation Study}
The excellent performance of EGR-Net comes from the double-branch scheme and the bridge connection between the two branches. In order to verify the effectiveness of the two key points in EGR-Net, the ablation analysis is performed in this subsection. First, only the bridge connection (BC) between the RSM-branch and the EGR-branch is removed, the first ablation model is obtained, that is called EGR-Net-No-BC. Then, the EGR-branch is completely deleted from EGR-Net, and a CNN model with only RSM as the input (CNN-RSM) is obtained. The compared results are provided in Table.~\ref{Tab:UCONN_AblationResults}. As shown in Table.~\ref{Tab:UCONN_AblationResults}, it is clear that the EGR-Net outperforms the EGR-Net-No-BC and the CNN-RSM under the four SNRs. The accuracy differences between the two ablated models and EGR-Net become more significant as the added noise power goes stronger. Compared with 97.09\% of EGR-Net, the EGR-Net-No-BC achieves 95.73\% accuracy under SNR=-6dB. So, the bridge connection contributes the 1.36\% improvement in accuracy. It means that the conversion of feature maps of the intermediate layers in RSM-branch can strengthen the feature learning ability of the CNN model. For the CNN-RSM, its average accuracy is only 86.15\% during SNR=-6dB, and our EGR-branch brings 9.58\% accuracy improvements. It can also be noticed that the standard deviation of EGR-Net decreases quicker than the EGR-Net-No-BC and the CNN-RSM, and it indicates that the EGR-branch can help improve the model stability.

\begin{figure}[!t]
  \centering
  \includegraphics[scale=0.25]{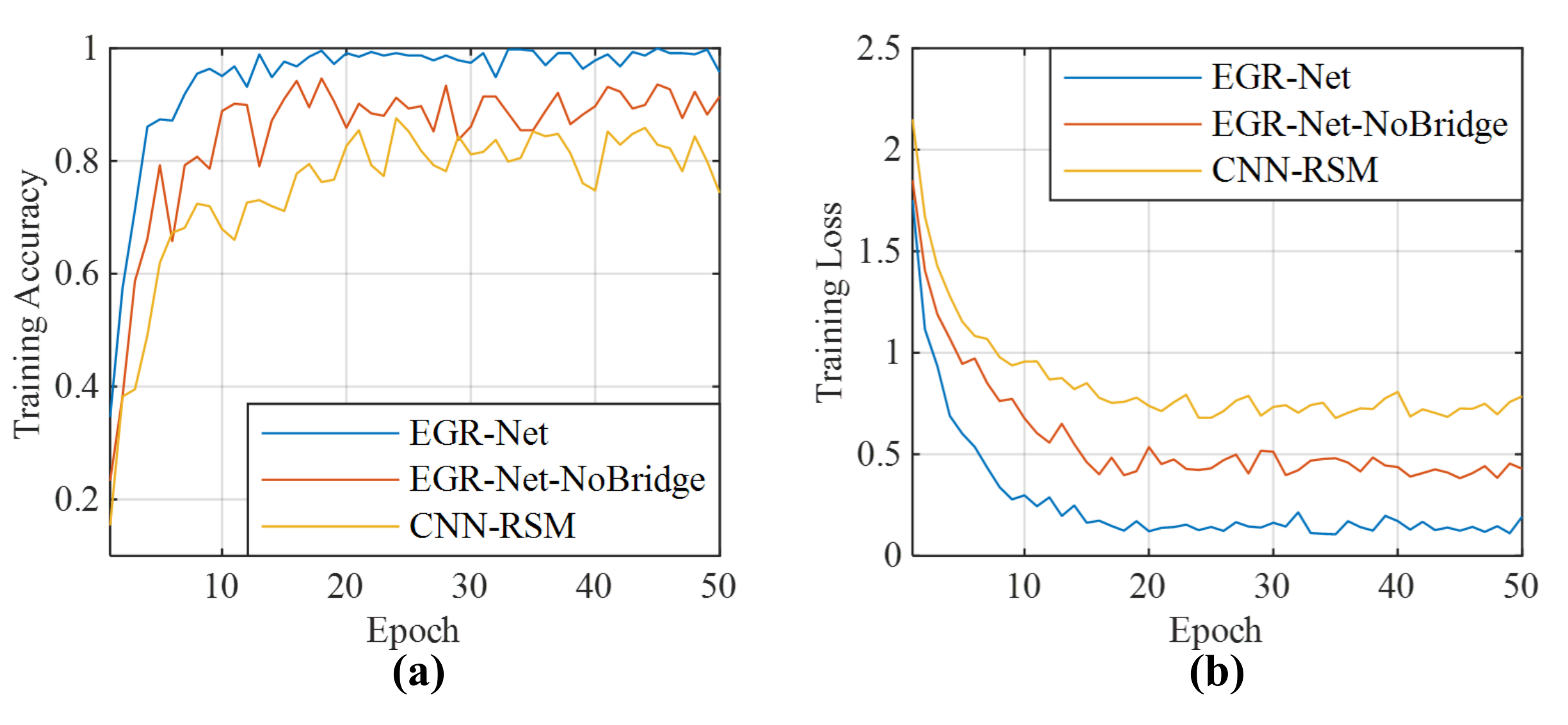}
  \caption{The training comparison of the EGR-Net, EGR-Net-No-BC, and CNN-RSM under SNR=-6dB. (a) training accuracy; (b) training loss.}
  \label{fig:UCONN_TRAIN_LOSS_ACC}
\end{figure}

Moreover, to better understand the convergence improvements brought by the double-branch scheme and bridge connection design in the training process, the training accuracy and CE loss curves of EGR-Net, EGR-Net-No-BC, and CNN-RSM in the first 50 epochs under SNR=-6dB are shown in Fig.~\ref{fig:UCONN_TRAIN_LOSS_ACC}. From the figure, it can be seen that the EGR-Net converges fastest in the three models. After ten epochs, the training accuracy of EGR-Net achieves around 95\%, and the training loss approaches 0.2, and the EGR-Net-No-BC reaches an accuracy of 90\% and 0.7 of the CE loss. As for the CNN-RSM model, it works worse than the other two models. Its final accuracy and CE loss after 50 epochs are only close to the corresponding curves of EGR-Net-No-BC at the 10-th epoch. Thus, the effectiveness of the two core novel points of EGR are verified.

\begin{table}[!t]
  \centering
  \caption{Comparison accuracy (\%) results of EGR-Net and CNN-RSM of the UCONN dataset under four SNR scenarios}
  \label{Tab:UCONN_AblationResults}
  \setlength{\tabcolsep}{1.5mm}{
  \begin{tabular}{lcccc}
    \toprule
    SNR & -6dB & -4dB & -2dB & 0dB \\
    \midrule
    EGR-Net & \textbf{97.09}±1.73 & \textbf{98.76}±0.82 & \textbf{99.70}±0.36 & \textbf{100.00}±0.00 \\
    EGR-Net-No-BC & 95.73±1.44  & 97.86±0.59 &99.44±0.24& 99.57±0.26\\
    CNN-RSM & 86.15±1.54 & 92.91±2.43 & 95.94±1.16 & 98.33±1.65 \\
    \bottomrule
  \end{tabular}
}
\end{table}

\subsubsection{Comparison with Other Methods}
In this part, the proposed EGR-Net is compared with six CNN-based state-of-the-art (SOTA) deep learning methods and three traditional shallow learning methods under four different noise levels (SNR=-6dB, -4dB, -2dB, and 0dB). For the six SOTA CNN methods, there are three 2D-CNN based models (Wen-CNN \cite{wen2017new} with RSM inputs, Shao-CNN \cite{shao2018highly} with the continuous wavelet transform inputs, Chen-CNN \cite{chen2020deep} using CSC as inputs) and three  1D-CNN based models (Zhang-CNN\cite{zhang2017new}, Zhao-CNN\cite{zhao2019deep}, Li-CNN\cite{li2021waveletkernelnet}). The selected three traditional methods are long short-term memory (LSTM), gated recurrent unit (GRU), and support vector machine (SVM) with input of selected sixteen manual features as given in \cite{pan2019novel}. For the LSTM and GRU methods, a single layer is used, and the number of units is 256. The average accuracies and standard deviations of each method after five runs are given in Table.~\ref{Tab:UCONN_COMPARATIVE_RESULTS}.

From the Table.~\ref{Tab:UCONN_COMPARATIVE_RESULTS}, it can be seen that the proposed EGR-Net shows the highest average accuracy and the most powerful noise robustness over the four SNR scenarios in the ten methods. First, among the three 2D-CNN methods, Wen-CNN\cite{wen2017new}, which takes the RSM as input, can achieve good accuracy but it suffers from a huge decrease of accuracy when SNR=-6dB. Although Wen-CNN\cite{wen2017new} has no information loss, the low separability of the original raw signal limits its performance. Chen-CNN\cite{chen2020deep}, whose input is CSC representation, obtains the worst results. Shao-CNN\cite{shao2018highly} is the best of the three CNN-based methods, working better than Wen-CNN\cite{wen2017new} and Chen-CNN\cite{chen2020deep}. It is because Shao-CNN\cite{shao2018highly} is based on transfer learning which allows them to adopt highly complicated models. Our proposed method can obtain higher classification accuracy than Shao-CNN\cite{shao2018highly}, although its pre-trained model VGG-16 contains more than 15 million trainable parameters, almost 35 times that of EGR-Net. And, the FLOPs of the VGG16 is 15.3 billion, which is much larger than the 1.26 billion of our proposed EGR-Net. It indicates that the proposed EGR-Net can extract discriminative features for recognizing faults more efficiently. And also, it can be noticed that the Shao-CNN\cite{shao2018highly} only achieves 85.35\% accuracy under SNR=-6dB, which is far lower than 97.09\% of EGR-Net. It means that the 2D representation of Shao-CNN is limited because of information loss. Our EGR-Net has the RSM branch to learn features from the original value, so such a problem can be avoided. In addition, Shao-CNN\cite{shao2018highly} and Chen-CNN\cite{chen2020deep} are not end-to-end models because their 1D-to-2D conversions are so complicated that they are difficult to be processed in the CNN model. Second, Li-CNN\cite{li2021waveletkernelnet} has the best diagnosis performance among the three 1D-CNN based methods due to its wavelet kernel. Zhang-CNN\cite{zhang2017new} achieives fair accuraies. Third, the three traditional fault diagnosis methods exhibit lower accuracies than the deep learning methods except for the Chen-CNN\cite{chen2020deep} and Zhao-CNN\cite{zhao2019deep}. In the three methods, SVM provides the highest accuracy on the gearbox diagnosis. Despite all these, the EGR-Net shows a 11.74\% improvement compared with Shao-CNN and 23.08\% improvement compared with SVM on the UCONN gearbox dataset when SNR=-6dB. Compared with other methods, EGR-Net exhibits a significant performance improvement under different SNRs. This enhancement is due to the highly discriminative features provided by EGR and less information loss.

\begin{table}[!t]
  \centering
  \caption{Comparison accuracy (\%) results on the UCONN dataset under four SNR scenarios}
  \label{Tab:UCONN_COMPARATIVE_RESULTS}
  \setlength{\tabcolsep}{2.0mm}{
  \begin{tabular}{lcccc}
    \toprule
    SNR & -6dB & -4dB & -2dB & 0dB \\
    \midrule
    EGR-Net & \textbf{97.09}±1.73 & \textbf{98.76}±0.82 & \textbf{99.70}±0.36 & \textbf{100.00}±0.00 \\
    Wen-CNN\cite{wen2017new} & 78.42±7.73 & 91.72±3.32 & 94.57±1.78 & 94.74±1.84 \\
    Shao-CNN\cite{shao2018highly} & 85.35±1.11 & 96.88±0.29 & 98.72±0.54 & 99.75±0.10 \\
    Chen-CNN\cite{chen2020deep} & 40.90±1.51 & 60.25±3.62 & 80.34±1.52 & 93.89±1.69 \\
    Zhang-CNN\cite{zhang2017new} &77.69±8.14&91.57±5.16& 96.67±2.09 & 99.40±0.39\\
    Zhao-CNN\cite{zhao2019deep} &18.58±7.49 &42.71±6.16 & 53.19±7.81 & 67.95±9.12\\
    Li-CNN\cite{li2021waveletkernelnet}  &95.38±2.38 &98.63±1.08 & 97.75±3.72 & 99.86±0.20\\
    LSTM \cite{pan2019novel}& 58.59±3.56 & 66.26±3.91 & 67.73±3.35 & 72.95±1.65 \\
    GRU \cite{pan2019novel}& 63.76±2.17 & 67.48±2.17 & 68.12±2.19 & 71.45±1.23 \\
    SVM \cite{pan2019novel}& 74.01±0.45 & 75.38±0.95 & 78.97±2.05 & 83.33±1.39 \\
    \bottomrule
  \end{tabular}
}
\end{table}

\subsection{Case Study 2: Fault Diagnosis of Bearings}
\subsubsection{Dataset Description}
The bearing dataset used for the experimental validation is from Case Western Reserve University (CWRU) bearing data center, which is a highly popular benchmark dataset in bearing fault diagnosis research \cite{smith2015rolling}. The motor bearings were given artificial faults with different diameters (i.e., 7mils, 14 mils, 21 mils) on the outer race, inner race, and ball, respectively. The motor was imposed loads of 0, 1, 2, and 3 hp under different rotating speeds. As depicted in Table.~\ref{Tab:CWRU_Dataset}, there are 10 typical health conditions, that is normal condition and three fault types under three fault severities. The vibrational data were recorded with a sampling frequency of 12kHz.

\begin{figure}[!t]
  \centering
  \includegraphics[scale=0.2]{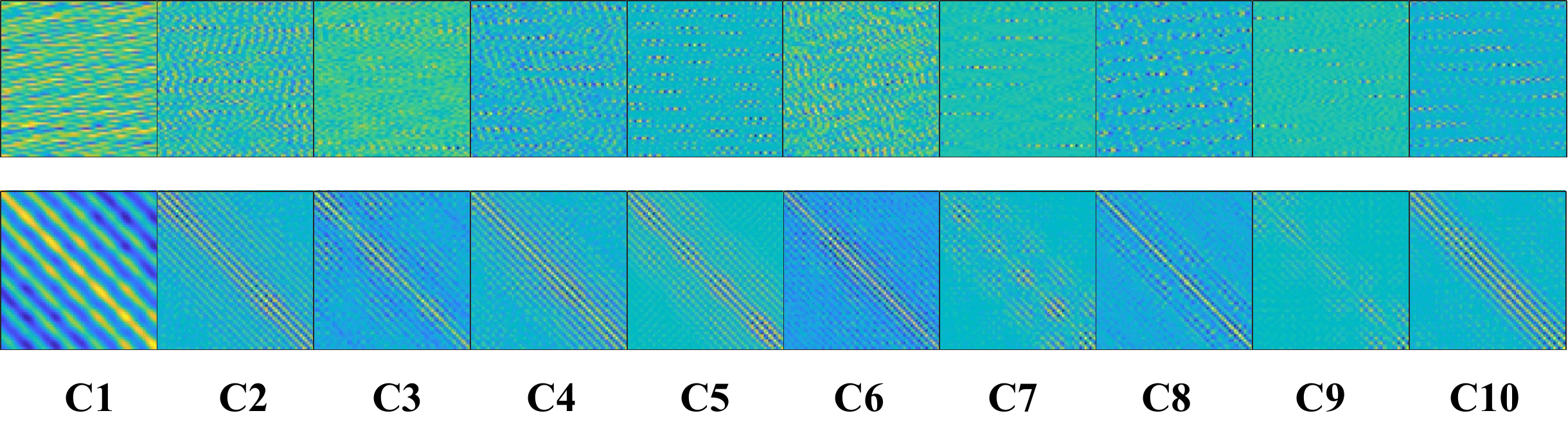}
  \caption{The RSMs (first row) and EGRs (second row) of ten health conditions of CWRU bearing dataset.}
  \label{fig:CWRU_RSM_EGR}
\end{figure}

In this paper, the vibrational signals of each health status were segmented into 800 samples with 4096 data points via the data augmentation with overlap strategy \cite{tang2020data}. The shape of RSM is set as  $64\times 64$, and the corresponding EGR has the same size as RSM. The first 50\% of the samples are for training and the rest 50\% are for testing. The raw signal matrices and EGRs of ten health conditions are given in Fig. \ref{fig:CWRU_RSM_EGR}. In those EGRs, the parallel stripe textures that indicate the periodic components of raw signals can be observed. The key information of faults is actually reflected in the periodic fluctuations of EGRs. Especially, the normal health condition shows an evenly distributed pattern, reflecting the intrinsic characteristic of the bearing vibration. Thus, the effective features related to faults are uncovered by EGRs.

\begin{table}[!t]
  \centering
  \caption{Description of ten bearing conditions of CWRU dataset.}
  \label{Tab:CWRU_Dataset}
  \begin{tabular}{llll}
  \toprule
      Bearing conditions & Fault diameter (mil) & Motor load (hp) & Label \\
  \midrule
      Normal     & Null & 0, 1, 2, 3 & C1 \\
      Ball       & 7    & 0, 1, 2, 3 & C2 \\
      Ball       & 14   & 0, 1, 2, 3 & C3 \\
      Ball       & 21   & 0, 1, 2, 3 & C4 \\
      Outer Race & 7    & 0, 1, 2, 3 & C5 \\
      Outer Race & 14   & 0, 1, 2, 3 & C6 \\
      Outer Race & 21   & 0, 1, 2, 3 & C7 \\
      Inner Race & 7    & 0, 1, 2, 3 & C8 \\
      Inner Race & 14   & 0, 1, 2, 3 & C9 \\
      Inner Race & 21   & 0, 1, 2, 3 & C10 \\
  \bottomrule
  \end{tabular}

\end{table}

\subsubsection{Texture Pattern of EGR on Bearing Fault Diagnosis}
\begin{figure}[!t]
  \centering
  \includegraphics[scale=0.2]{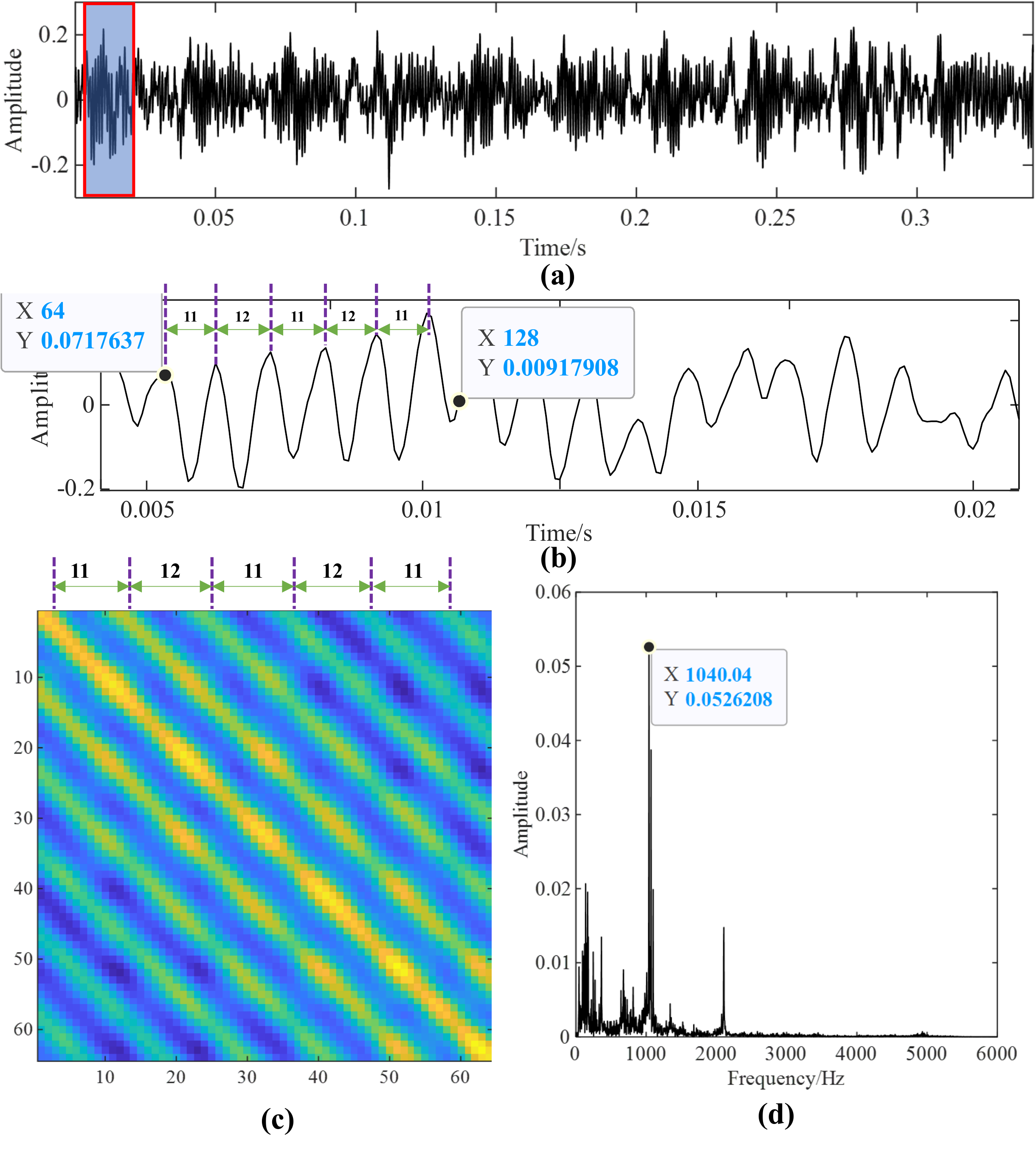}
  \caption{The illustration of the EGR principle using the "Normal-0" file in CWRU bearing dataset. (a) The waveform of the signal; (b) The local views of the first 300 data points; (c) The EGR of the signal; (d) The FFT of the signal.}
  \label{fig:CWRU_SIGNAL_EGR_FFT}
\end{figure}

As mentioned earlier, EGR can capture the periodic fluctuation related to faults in the vibrational signals, and the periodicity can be reflected in the textures of EGR. In this part, we will analyze the relationship between the periodic component of vibrational signals and EGR stripe texture and explore deeper characteristics of EGR. For the reproduction by others, the first 4096 data points of the "Normal-0" file in the CWRU dataset is used, and the signal segment is shown in Fig.~\ref{fig:CWRU_SIGNAL_EGR_FFT}(a).
For the convenience of the display, we only concentrate on the first 300 data points. The local view of the signal segment is shown in Fig.~\ref{fig:CWRU_SIGNAL_EGR_FFT}(b). From the subfigure, it can be noticed that there is a sine-line wave with a period of 11 or 12 sampling points. The EGR of the signal is given in Fig.~\ref{fig:CWRU_SIGNAL_EGR_FFT}(c), and the EGR is a symmetric matrix with fixed period stripes. The period of the stripes in EGR is the same as the raw signal in Fig.~\ref{fig:CWRU_SIGNAL_EGR_FFT}(b). And also, it can be seen that the signal has this fixed period in all of the data points in the data file. The signal's amplitude fluctuations are just amplitude modulation with the same carrier signal. Thus, the EGRs of all samples in this working condition have similar but different texture patterns.To better understand the principle of EGR, the FFT spectrum of the signal is provided in Fig.~\ref{fig:CWRU_SIGNAL_EGR_FFT}(d). It can be observed that the highest spectrum peak is at around 1040 Hz, and this actually represents the intrinsic period of the signal, that is ${12000}/{1040}=11.5385$ sample points in average, which is very close to the abovementioned period of 11 or 12 data points in EGR. It can be concluded that most unimportant details are discarded by EGR, the useful information are projected into the texture changes in the representation. Compared with classifying the 1D time-varying vibrational signals with complicate modulations directly, distinguishing the EGR images with simple textures is not such a hard work.

\subsubsection{Model Visualization Analysis}

\begin{figure*}[!t]
  \centering
  \includegraphics[scale=0.18]{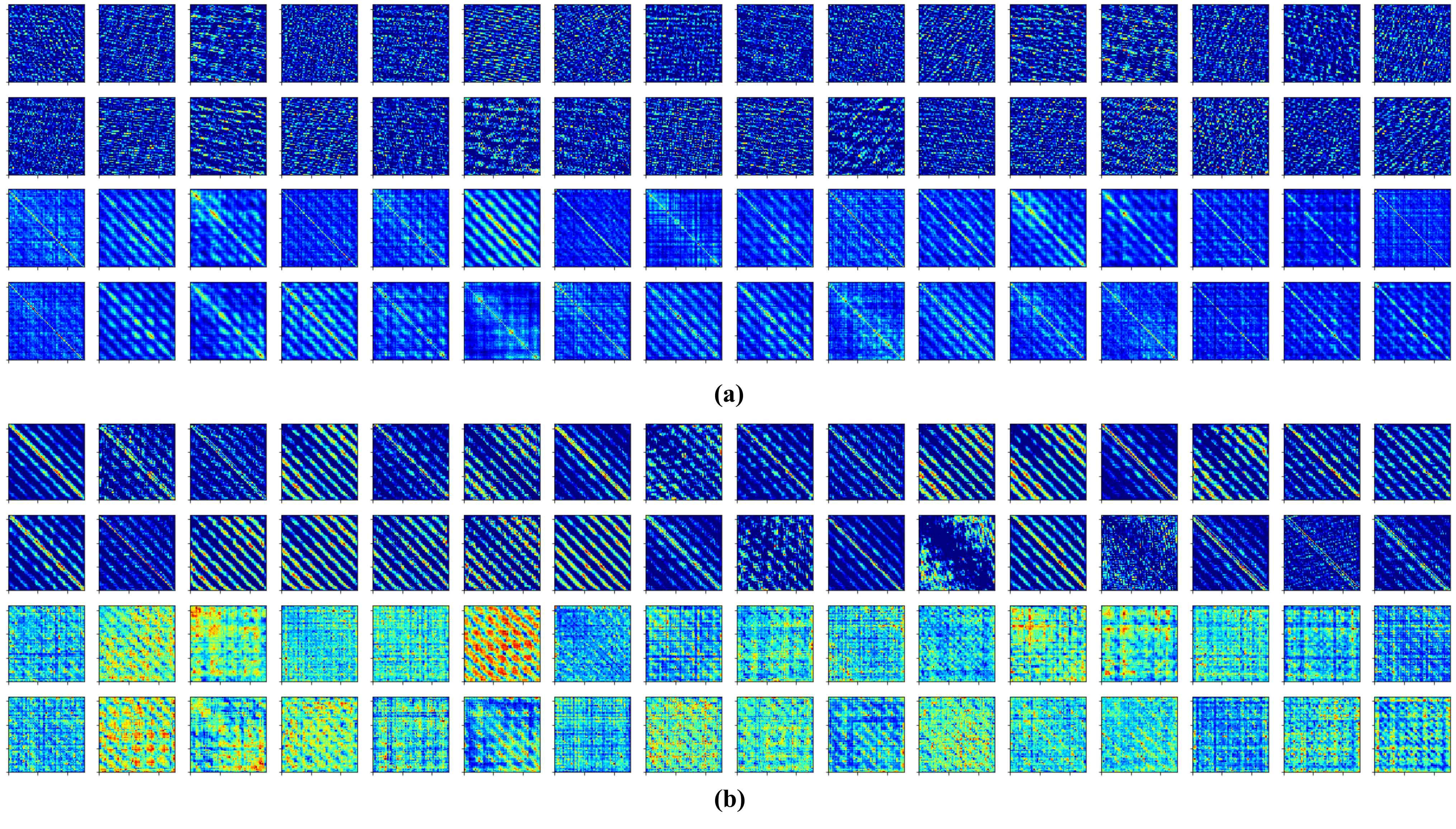}
  \caption{The outputs of the first GCB in EGR-Net (SNR=0dB). (a) The output of the RSM branch of the first GCB and the EGR of the RSM branch output. The top two rows are corresponding to the ${{X}_{h}}$ in Eq. (\ref{Eq:04}), and the bottom two rows are the $\text{Gram}({{X}_{h}})$ in Eq. (\ref{Eq:05}); (b) The output of the EGR-branch of the first GCB. The top two rows are ${{G}_{h}}$ in Eq (\ref{Eq:05}), and the third and fourth rows of (b) are $X{{G}_{h}}$ in Eq. (\ref{Eq:06}).}
  \label{fig:CWRU_INTERMEDIATE_LAYER}
\end{figure*}

To better understand the feature learning mechanism of the proposed EGR-Net, we visualize the feature maps of the first GCB block, as shown in Fig. \ref{fig:CWRU_INTERMEDIATE_LAYER}. The 32-channel output of the RSM branch are illustrated in the top two rows in Fig. \ref{fig:CWRU_INTERMEDIATE_LAYER}(a). We can notice that the textures of RSM feature maps are not significantly distinguishable intuitively, although the RSM contains the richest original information. The EGRs of the RSM output feature maps are shown in the bottom two rows in Fig. \ref{fig:CWRU_INTERMEDIATE_LAYER}(a). It can be seen that the unregular RSM feature maps in the top two rows become much clear when they are converted into EGRs. The converted EGR provides additional information and enhance the feature diversity for the CNN model. Compared with the RSM representation, the EGR always has simpler textures, which is beneficial for the feature learning of the CNN model. The 64-channel feature maps of the EGR branch are depicted in Fig. \ref{fig:CWRU_INTERMEDIATE_LAYER}(b). It also includes two parts, the first part (top two rows) contains 32-channel feature maps of EGR of the raw signals, and the second part (bottom two rows) contains 32-channel LN-processed EGRs of the 32-channel RSM feature maps. It can be observed that different channels in the top two rows focus on different texture features of EGRs. For example, the 7-th channel tends to extract the stripes' features in the middle, but the 6-th channel learns more features from the stripes far away from the diagonal. As for the second part, the EGRs after LN operation show the different textures patterns. But the stripe textures are still distinguishable.
Our proposed EGR-Net can handle raw vibrational signals, and its EGRs simultaneously. EGR-Net can reduce the information loss due to 1D-to-2D conversion than the methods using only one representation. This scheme can be done because of the easy calculation of EGR via matrix multiplication.

\subsubsection{Comparison with Other Methods}
Similar to the experiments of gear diagnosis in the first case study, the proposed EGR-Net is compared with the sample six methods under SNR=-6dB, -4dB, -2dB, and 0dB, respectively. Each situation is run with five trials to reduce randomness, and the mean value and standard deviation of the accuracy are listed in Table.~\ref{Tab:CWRU_COMPARATIVE_RESULTS}. As we can see, the EGR-Net outperforms all other methods. The EGR-Net reaches 99.05\% accuracy even under SNR=-6dB. Among the nine comparative methods, the accuracies of Shao-CNN\cite{shao2018highly} are higher than the other nine methods at the cost of a heavy computation burden. However, there is still a large performance gap between Shao-CNN\cite{shao2018highly} and our EGR-Net in the average accuracy. Similar to the gear diagnosis, the Wen-CNN\cite{wen2017new} still works badly on strong background noise, and Chen-CNN works the worst in all three CNN-based methods because of the limited model capacity. Li-CNN\cite{li2021waveletkernelnet} shows the weakest performance in the three 1D-CNN methods. GRU is the best of the three traditional methods, and it is better than Wen-CNN when SNR = -6dB and -4dB. The comparison results further demonstrate the superiority of the proposed EGR-Net.

\begin{table}[!t]
  \centering
  \caption{Comparison accuracy (\%) results on the CWRU dataset under four SNR scenarios}
  \label{Tab:CWRU_COMPARATIVE_RESULTS}
  \setlength{\tabcolsep}{2.0mm}{
  \begin{tabular}{lcccc}
    \toprule
    SNR & -6dB & -4dB & -2dB & 0dB \\
    \midrule
    EGR-Net & \textbf{99.05}±0.10 & \textbf{99.73}±0.05 & \textbf{99.91}±0.05 & \textbf{99.96}±0.03 \\
    Wen-CNN\cite{wen2017new} & 76.35±3.62 & 91.03±2.15 & 96.01±1.14 & 97.56±0.57 \\
    Shao-CNN\cite{shao2018highly} & 96.53±0.10 & 98.53±0.07 & 99.08±0.18 & 99.42±0.08 \\
    Chen-CNN\cite{chen2020deep} & 57.15±1.60 & 69.88±1.58 & 78.92±0.98 & 85.97±1.08 \\
    Zhang-CNN\cite{zhang2017new} &91.58±8.26&95.55±5.61& 94.96±5.14 & 98.48±1.22\\
    Zhao-CNN\cite{zhao2019deep} &86.56±6.94&91.36±3.27& 93.65±2.54 & 97.00±0.82\\
    Li-CNN\cite{li2021waveletkernelnet} &75.31±1.82&86.56±3.01& 93.41±2.66 & 96.17±1.10\\
    LSTM\cite{pan2019novel} & 89.63±4.18 & 92.40±3.95 & 94.34±1.24 & 95.83±1.05 \\
    GRU\cite{pan2019novel} & 94.97±0.54 & 96.41±0.41 & 97.46±0.36 & 97.83±0.21 \\
    SVM\cite{pan2019novel} & 85.81±0.29 & 90.35±0.38 & 92.81±0.15 & 93.73±0.80 \\
    \bottomrule
  \end{tabular}
}
\end{table}

\section{Conclusion}\label{sec:conclusion}
This paper introduced a new 1D to 2D conversion method called EGR to address the problems of complicated computation and low separability of the current 2D representation methods. The proposed EGR can capture the periodic components in vibrational signals and extract essential separable features. The EGR and RSM are fed into a double-branch CNN model called EGR-Net to reduce the information loss resulting from the 2D representation conversion and improve the noise robustness. The EGR-Net can make fault diagnoses in an end-to-end manner. The performance of EGR-Net was verified on two widely studied bearing and gearbox datasets. The results show that the proposed method can outperform the state-of-the-art deep learning models and traditional intelligent methods. In the future work, we will explore the characteristics of EGR and EGR-Net in remaining useful life problems.



\bibliographystyle{elsarticle-num}
\bibliography{my_references}





\end{document}